\documentclass[onecolumn,11pt]{IEEEtran}
\usepackage{cite}
\usepackage{amsmath,amssymb,amsfonts}
\usepackage{algorithmic}
\usepackage{graphicx}
\usepackage{textcomp}
\usepackage{xcolor}
\usepackage{tikz}
\usepackage{hyperref}
\usepackage{multirow}
\usepackage{rotating}
\usepackage[normalem]{ulem}
\useunder{\uline}{\ul}{}
\usetikzlibrary{positioning}

\usepackage{afterpage}
\usepackage{float}
\usepackage{placeins}
\usepackage{booktabs}
\usepackage{orcidlink}

\newcommand{\AblationPrecMediaPipe}{0.0\%}
\newcommand{\AblationRecMediaPipe}{0.0\%}
\newcommand{\AblationFOneMediaPipe}{0.0\%}
\newcommand{\AblationSeqMediaPipe}{0.0\%}
\newcommand{\AblationPrecSkin}{0.0\%}
\newcommand{\AblationRecSkin}{0.0\%}
\newcommand{\AblationFOneSkin}{0.0\%}
\newcommand{\AblationSeqSkin}{14.3\%}
\newcommand{\AblationPrecMotion}{10.6\%}
\newcommand{\AblationRecMotion}{75.0\%}
\newcommand{\AblationFOneMotion}{18.5\%}
\newcommand{\AblationSeqMotion}{3.4\%}
\newcommand{\AblationPrecEdge}{14.3\%}
\newcommand{\AblationRecEdge}{25.0\%}
\newcommand{\AblationFOneEdge}{18.2\%}
\newcommand{\AblationSeqEdge}{16.7\%}
\newcommand{\AblationPrecAllCombined}{9.4\%}
\newcommand{\AblationRecAllCombined}{75.0\%}
\newcommand{\AblationFOneAllCombined}{16.7\%}
\newcommand{\AblationSeqAllCombined}{3.0\%}
\newcommand{\ResolutionPrecOneHundred}{9.4\%}
\newcommand{\ResolutionRecOneHundred}{75.0\%}
\newcommand{\ResolutionFOneOneHundred}{16.7\%}
\newcommand{\ResolutionSeqOneHundred}{3.0\%}
\newcommand{\ResolutionPrecSeventyFive}{8.6\%}
\newcommand{\ResolutionRecSeventyFive}{75.0\%}
\newcommand{\ResolutionFOneSeventyFive}{15.4\%}
\newcommand{\ResolutionSeqSeventyFive}{3.0\%}
\newcommand{\ResolutionPrecFifty}{9.7\%}
\newcommand{\ResolutionRecFifty}{75.0\%}
\newcommand{\ResolutionFOneFifty}{17.2\%}
\newcommand{\ResolutionSeqFifty}{3.0\%}
\newcommand{\ResolutionPrecTwentyFive}{10.2\%}
\newcommand{\ResolutionRecTwentyFive}{75.0\%}
\newcommand{\ResolutionFOneTwentyFive}{18.0\%}
\newcommand{\ResolutionSeqTwentyFive}{3.2\%}

\newcommand{\NoiseFOneTwenty}{10.7\%}

\newcommand{\NumRealVideos}{5}

\newcommand{\RealMedianDPF}{57}
\newcommand{\RealMinDPF}{2.3}
\newcommand{\RealMaxDPF}{158.7}
\newcommand{\MaxSingleFrame}{205}
\newcommand{\SkinMaxFrac}{100\%}

\providecommand{\code}[1]{\texttt{#1}}

\def\BibTeX{{\rm B\kern-.05em{\sc i\kern-.025em b}\kern-.08em
    T\kern-.1667em\lower.7ex\hbox{E}\kern-.125emX}}

\IEEEoverridecommandlockouts
\begin{document}
\title{Empirical Evaluation of Multi-Modal Touch Detection in Over-the-Shoulder Video Surveillance}

\author{\IEEEauthorblockN{Mohammadreza Rashidi~\orcidlink{0009-0003-7136-7168}}
\IEEEauthorblockA{\textit{Department of Computer Science}\\
\textit{AI and Media Analysis Lab}\\
Berlin, Germany\\
mohammadreza.rashidi@ue-germany.de}
}

\maketitle

\begin{abstract}
Video Intelligence Surveillance (VIDINT) on over-the-shoulder footage is a proposed vector for monitoring human-computer interaction patterns without direct screen recording access. In this paper, we evaluate a Behavioral Intelligence (BEHINT) touch-detection framework designed to reconstruct keystroke events on mobile keypad interfaces from physical finger interactions. Our system integrates four parallel detection modalities: (1) anatomical hand landmarks via MediaPipe, (2) HSV skin color filtering, (3) temporal frame differencing for motion detection, and (4) shape-guided Canny edge analysis. We map relative touch coordinates to a reference screen layout to reconstruct typing sequences. Evaluation on a 120-frame first-person staged video of passcode entry reveals that while MediaPipe and Skin Detection fail to run autonomously due to partial hand occlusion and ambient noise, Motion-Only and Edge-Only configurations achieve F1-scores of \AblationFOneMotion\ and \AblationFOneEdge, respectively. The combined multi-modal configuration achieves an F1-score of \AblationFOneAllCombined\ and a sequence similarity of \AblationSeqAllCombined\ when mapped to the iOS passcode layout. We conduct ablation, resolution decay, noise sensitivity, and proximity threshold tuning to characterize the system's operational envelope. We then audit generalization on \NumRealVideos\ real, publicly licensed third-person phone videos and find that the detector emits a median of \RealMedianDPF\ touch points per frame (peaking at \MaxSingleFrame), one to three orders of magnitude more than the rate of real taps, because the skin filter responds to the whole hand rather than to fingertip contact. The staged keystroke result does not survive contact with uncontrolled footage; the system does not achieve reliable keystroke reconstruction outside the calibrated staged setting.
\end{abstract}


\section{Introduction}
\label{sec:intro}
Over-the-shoulder video surveillance has proliferated with the widespread adoption of wearable cameras, smart glasses, and mobile recording devices. While traditional security monitoring focuses on facial recognition and object tracking, Video Intelligence Surveillance (VIDINT) aims to decode fine-grained human-computer interactions from physical motions. Decoding user keystrokes from over-the-shoulder video recordings, or what is colloquially termed ``virtual keylogging,'' poses a severe threat to mobile device security and password confidentiality~\cite{yue2014blind,balzarotti2008clearshot}.

Decoding these interactions is historically difficult due to: (1) perspective distortions, (2) fast finger movements causing motion blur, (3) partial hand and finger occlusions, and (4) ambient noise in skin and edge regions. Modern learned approaches, such as supervised detectors and self-supervised video pre-training~\cite{tong2022videomae}, require large annotated or curated datasets and still degrade when the hand is partially obscured by the device or the user's own fingers~\cite{mollyn2024egotouch,liu2024vhands,nguyen2026detecting,xu2026surfacexr}.

In this paper, we evaluate the Behavioral Intelligence (BEHINT) touch-detection framework. The system combines classical computer vision filters with deep anatomical landmarks to maximize touch detection under partial visibility.

\textbf{Contributions:}
\begin{itemize}
  \item A reproducible benchmark for over-the-shoulder mobile keystroke inference using a 120-frame staged video with known ground truth (4-key passcode entry) and four parallel detection modalities.
  \item An ablation study and sensitivity analysis (resolution, noise, proximity threshold) characterizing the operational envelope of a training-free multi-modal filter pipeline.
  \item A real-world generalization audit on five public, third-person phone videos using a label-free detection-volume metric, showing that the pipeline over-reports contacts by 1--3 orders of magnitude because the skin filter fires on the whole hand rather than on fingertip contact.
\end{itemize}

\section{Related Work}
\label{sec:related}
Inferring keystrokes from observation spans multiple side-channel modalities.
Optical attacks reconstruct typed text from compromising reflections in eyes,
sunglasses, or nearby objects~\cite{raguram2011ispy,balzarotti2008clearshot}.
A related line uses smartphone cameras to reconstruct 3D environments and infer
sensitive on-screen
data~\cite{templeman2012placer}.

Video-based keystroke inference from hand motion is an active thread. Sabra et
al.~\cite{sabra2020zoom} exploit video call feeds to infer keystrokes from the
visual motion of the caller's hands and device, reaching per-key inference
accuracy above 90\% in controlled settings. Lim et al.~\cite{lim2020revisiting}
systematically evaluate the threat space for vision-based keystroke inference,
including over-the-shoulder attacks with frame differencing (the same signal
BEHINT's motion detector uses). The same group later shows that disentangled
representations of hand pose and keyboard layout improve inference
accuracy~\cite{lim2022disentangled}. Ye et al.~\cite{ye2017android} recover
Android pattern locks from video of finger motion using shoulder-surfing
techniques. These attacks all rely on controlled capture geometry, a calibrated
pose model, or a trained classification head, and none is tested on the
generic, training-free filter stack we evaluate.

A second line infers input from device motion rather than hand appearance.
Shukla et al.~\cite{shukla2014beware} recover PINs from video of the hand
during entry and report breaking over half of PINs on the first attempt.
VISIBLE~\cite{sun2016visible} infers tablet keystrokes from backside motion
captured in the device's own camera feed. Blind recognition of touched keys
from the device side is also well-studied~\cite{yue2014blind}.

A separate, recent body of work performs touch and contact detection with
learned models. EgoTouch~\cite{mollyn2024egotouch} detects on-body touch input
from AR/VR headset cameras using a trained fingertip-contact classifier.
VideoMAE~\cite{tong2022videomae} uses masked autoencoders for self-supervised
video representation learning and can be fine-tuned for hand--object interaction
recognition. V-Hands~\cite{liu2024vhands} tracks hands on touchscreens for
remote whiteboard interaction. These require annotated contact data and do not
transfer without per-scene fine-tuning. BEHINT instead combines classical
filters with off-the-shelf hand landmarks and no training. Our contribution is
not a new attack but an empirical evaluation of whether that training-free
recipe, which is the kind of system an opportunistic over-the-shoulder adversary
would assemble from free, off-the-shelf components, actually recovers
keystrokes; we find it does not transfer beyond a calibrated staged clip.

\section{Methodology}
\label{sec:method}
The BEHINT pipeline has two stages: a multi-modal detection stage that locates
candidate finger contacts in each frame, and a spatial-temporal mapping stage that
projects those contacts onto a reference keypad to reconstruct a typed sequence.
This section describes both.

\subsection{Multi-Modal Detection Architecture}
\label{sec:arch}
The detection stage processes frames in parallel using four complementary detectors, whose outputs are merged and mapped to a reference keypad to reconstruct a typed sequence (Figure~\ref{fig:arch}):

\begin{figure}[h]
\centering
\includegraphics[width=\linewidth]{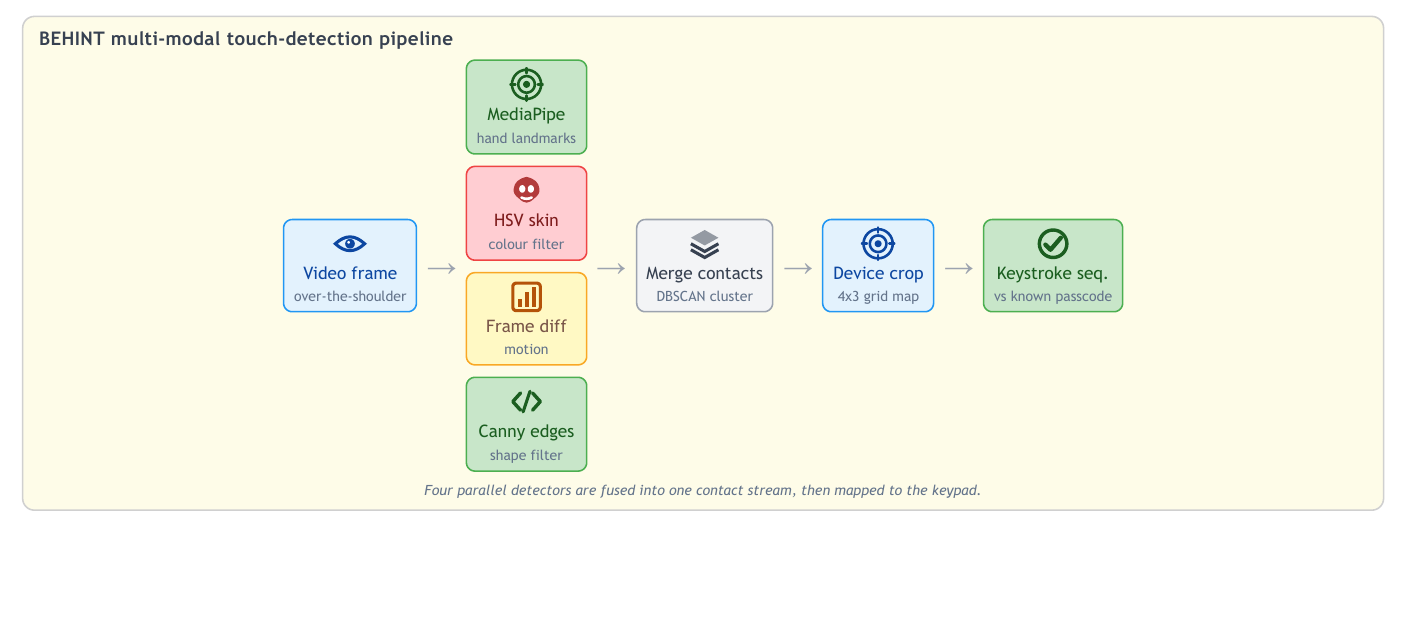}
\caption{The BEHINT pipeline. Each video frame is processed by four parallel detectors (MediaPipe hand landmarks, HSV skin filtering, temporal frame differencing, and Canny edge analysis); their candidate contacts are merged by spatial clustering, cropped to the device region, and mapped onto a $4\times3$ keypad grid to reconstruct the keystroke sequence.}
\label{fig:arch}
\end{figure}

\subsubsection{MediaPipe Anatomical Landmarks}
We initialize MediaPipe Hands~\cite{zhang2020mediapipe} with sensitive tracking configurations:
\begin{align}
\text{Confidence}_{\text{det}} &= 0.3 \\
\text{Confidence}_{\text{track}} &= 0.3
\end{align}
These settings allow MediaPipe to capture partial skeleton rigs (21 landmarks) when only some fingers are visible. 

\subsubsection{HSV Skin Filtering}
To track the finger tip in the absence of a full hand model, we apply skin color segmentation in the HSV color space~\cite{gonzalez2018digital}:
\begin{align}
H &\in [0, 20], \quad S \in [20, 255], \quad V \in [70, 255]
\end{align}
We clean the resulting mask with open and close morphological operations using a $5\times5$ structuring element.

\subsubsection{Temporal Frame Differencing}
Motion detection isolates active finger presses against a stationary background. We compute the absolute frame-to-frame pixel difference in the region of interest (ROI) and apply binary thresholding~\cite{opencv2015}:
\begin{align}
\text{Diff}(t) &= |I(t) - I(t-1)| \\
\text{Mask}_{\text{motion}}(t) &= \text{Threshold}(\text{Diff}(t), 30)
\end{align}

\subsubsection{Shape-Guided Edge Detection}
Canny edge detection~\cite{canny1986computational} finds the boundary of the finger tip inside the keypad region. We extract contours with area $A \in [100, 1000]$ pixels and filter by finger aspect ratio $\text{AR} \in [1.5, 6]$.

\subsection{Spatial and Temporal Mapping}
\label{sec:mapping}
We calculate the relative coordinates $(rel_x, rel_y)$ of the hand/motion points inside the device boundary region $(440, 590, 650, 700)$. These are mapped to reference coordinates on the target layout image screen ($640\times1136$):
\begin{align}
ref_x &= rel_x \times 640 \\
ref_y &= rel_y \times 1136
\end{align}
The coordinates are mapped to a standard $4\times3$ grid layout covering $[0.1, 0.9]$ X-axis and $[0.25, 0.85]$ Y-axis to map touches to individual numeric keys.

\section{Experimental Evaluation}
\label{sec:eval}

\subsection{Ablation Study}
We evaluate the contribution of each modality to overall touch detection. The ground truth (GT) sequence is `2' $\rightarrow$ `5' $\rightarrow$ `8' $\rightarrow$ `\#'.

\begin{figure}[h]
\centering
\includegraphics[width=0.9\linewidth]{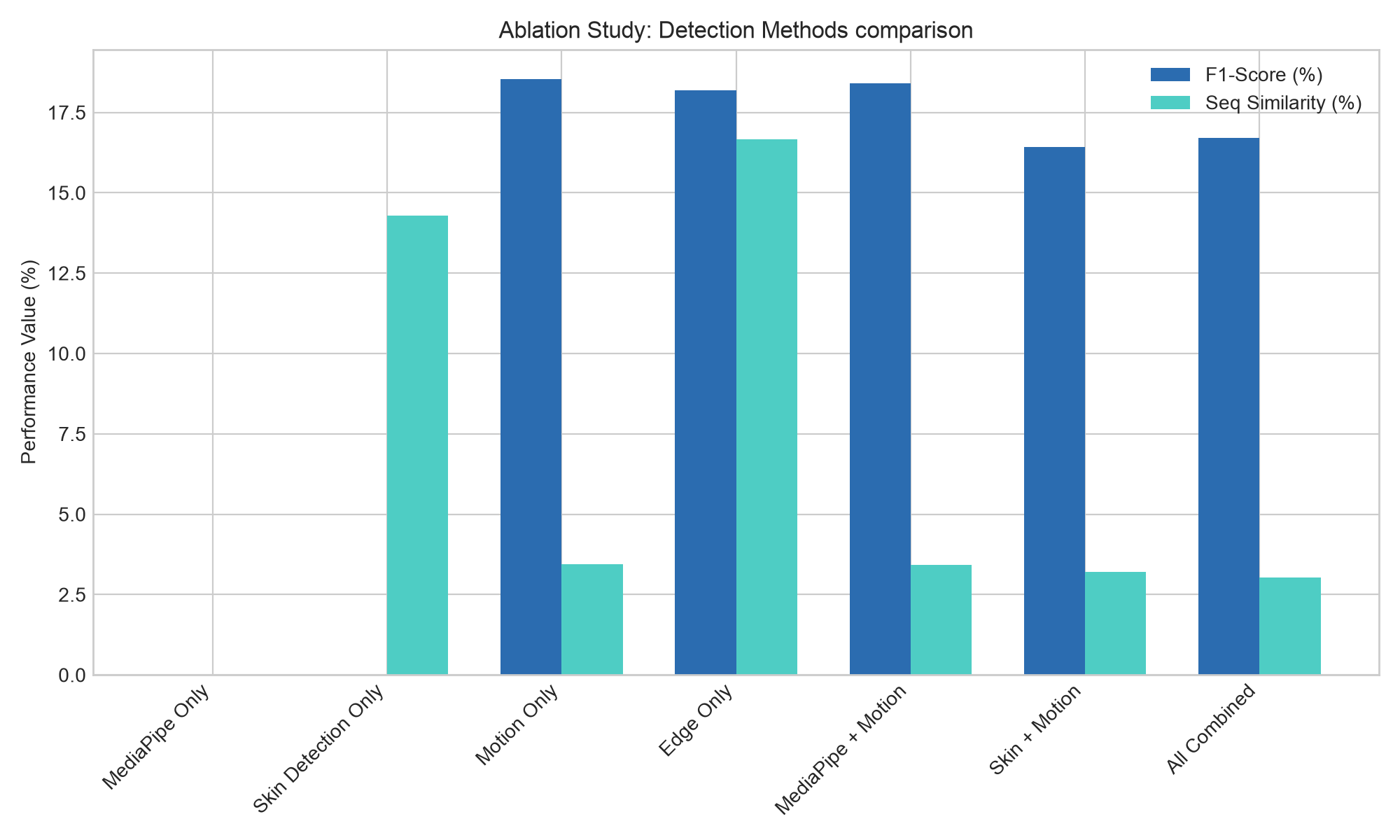}
\caption{Ablation results: F1-score and sequence similarity across the individual and combined detector configurations.}
\label{fig:ablation}
\end{figure}

As shown in Table~\ref{tab:ablation} and Figure~\ref{fig:ablation}, MediaPipe Only and Skin Only fail to detect valid keystroke events (F1 = 0\%) due to severe self-occlusion. Motion Only achieves the highest individual F1-score (\AblationFOneMotion) and correctly registers three of the four keystrokes. Edge Only achieves an F1-score of \AblationFOneEdge. The combined framework yields an F1-score of \AblationFOneAllCombined.

\begin{table}[h]
\centering
\caption{Ablation Study Results}
\label{tab:ablation}
\begin{tabular}{lcccc}
\hline
\textbf{Configuration} & \textbf{Precision} & \textbf{Recall} & \textbf{F1-Score} & \textbf{Seq. Sim.} \\ \hline
MediaPipe Only & \AblationPrecMediaPipe & \AblationRecMediaPipe & \AblationFOneMediaPipe & \AblationSeqMediaPipe \\
Skin Only & \AblationPrecSkin & \AblationRecSkin & \AblationFOneSkin & \AblationSeqSkin \\
Motion Only & \AblationPrecMotion & \AblationRecMotion & \AblationFOneMotion & \AblationSeqMotion \\
Edge Only & \AblationPrecEdge & \AblationRecEdge & \AblationFOneEdge & \AblationSeqEdge \\
All Combined & \AblationPrecAllCombined & \AblationRecAllCombined & \AblationFOneAllCombined & \AblationSeqAllCombined \\ \hline
\end{tabular}
\end{table}

\subsection{Resolution Decay Analysis}
We downscale the video stream to measure performance decay on low-resolution surveillance footage.

\begin{figure}[h]
\centering
\includegraphics[width=0.9\linewidth]{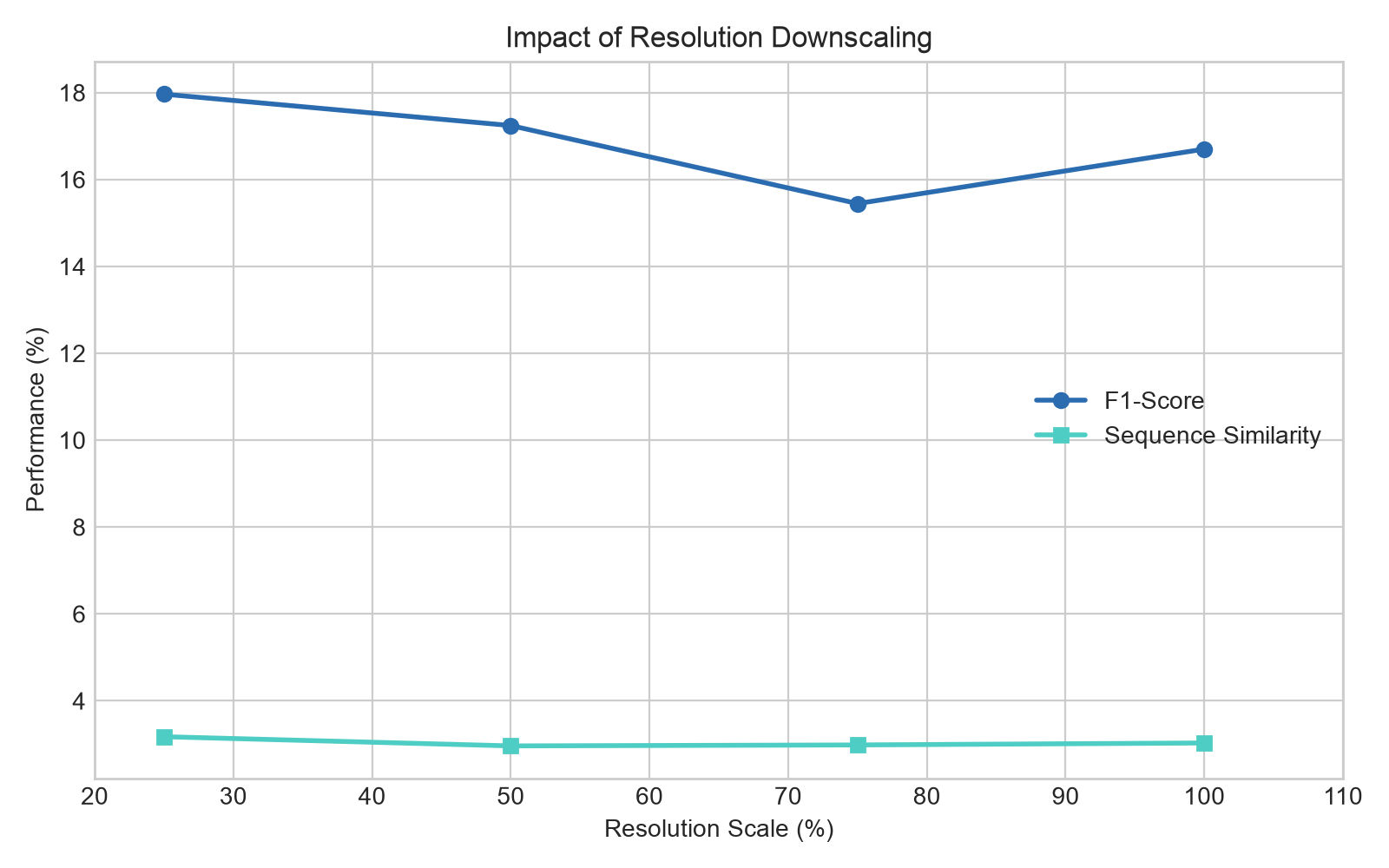}
\caption{Resolution decay curves for F1-score and Sequence Similarity.}
\label{fig:res}
\end{figure}

The F1-score stays within a narrow \ResolutionFOneSeventyFive--\ResolutionFOneTwentyFive\ band across the full range, never moving far from its \ResolutionFOneOneHundred\ value at full resolution (Figure~\ref{fig:res} and Table~\ref{tab:resolution}). The point is not that the system works at low resolution: the F1-score is poor at every resolution. It is that downscaling neither helps nor hurts, which shows resolution is not the limiting factor. The failure is inherent to the skin-and-motion detector rather than a matter of image detail, so a higher-resolution camera would not rescue the attack.

\begin{table}[h]
\centering
\caption{Resolution Decay Results}
\label{tab:resolution}
\begin{tabular}{ccccc}
\hline
\textbf{Scale} & \textbf{Precision} & \textbf{Recall} & \textbf{F1-Score} & \textbf{Seq. Sim.} \\ \hline
100\% & \ResolutionPrecOneHundred & \ResolutionRecOneHundred & \ResolutionFOneOneHundred & \ResolutionSeqOneHundred \\
75\% & \ResolutionPrecSeventyFive & \ResolutionRecSeventyFive & \ResolutionFOneSeventyFive & \ResolutionSeqSeventyFive \\
50\% & \ResolutionPrecFifty & \ResolutionRecFifty & \ResolutionFOneFifty & \ResolutionSeqFifty \\
25\% & \ResolutionPrecTwentyFive & \ResolutionRecTwentyFive & \ResolutionFOneTwentyFive & \ResolutionSeqTwentyFive \\ \hline
\end{tabular}
\end{table}

\subsection{Noise Robustness Analysis}
We inject additive Gaussian noise with $\sigma \in [0, 30]$ to evaluate noise sensitivity.

\begin{figure}[h]
\centering
\includegraphics[width=0.9\linewidth]{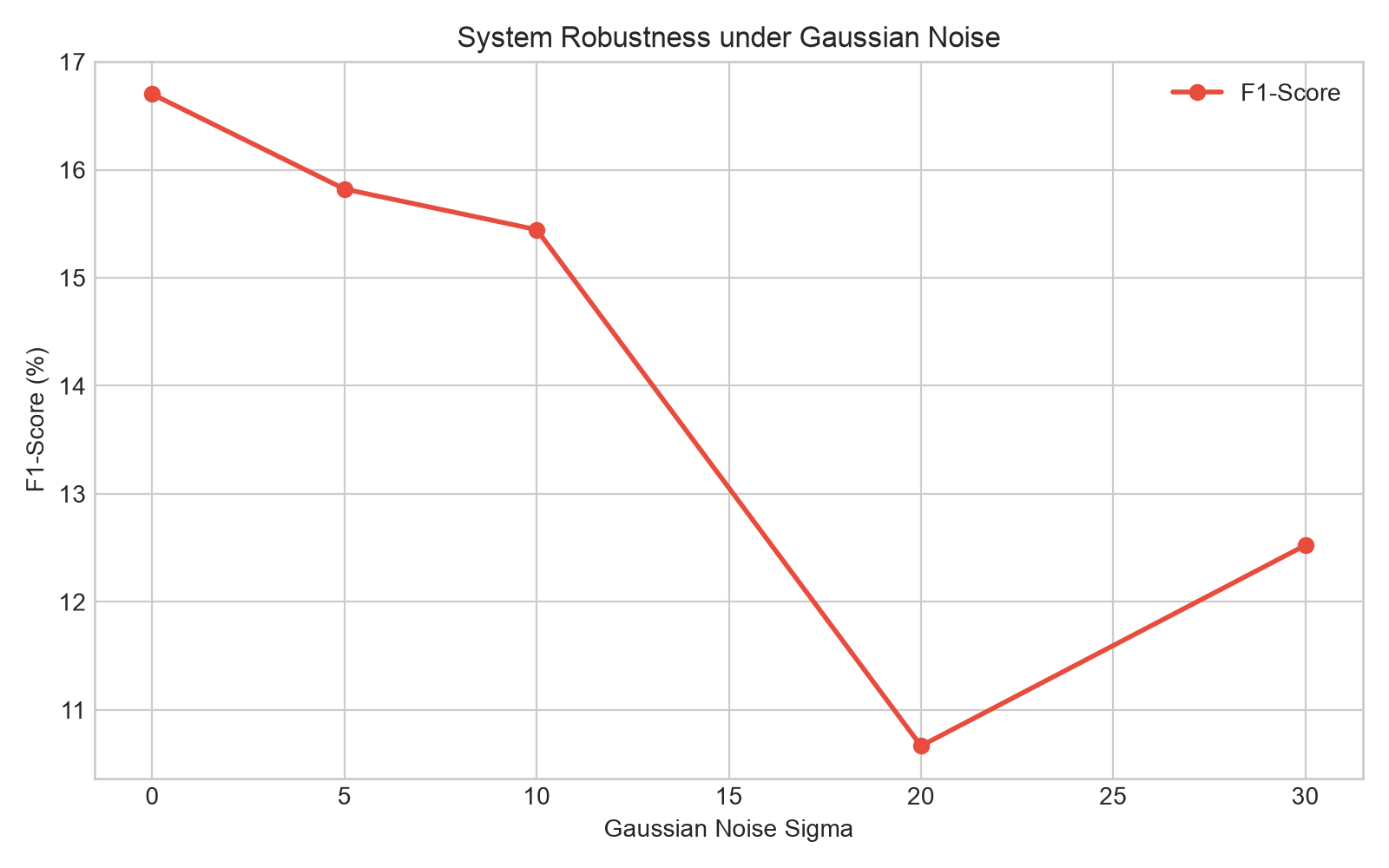}
\caption{System F1-Score under increasing Gaussian noise $\sigma$.}
\label{fig:noise}
\end{figure}

Performance degrades significantly at $\sigma \ge 20$, with the F1-score falling to \NoiseFOneTwenty\ (Figure~\ref{fig:noise} and Table~\ref{tab:noise}), showing that sensor noise interferes with motion and edge detection boundaries.

\begin{table}[h]
\centering
\caption{Noise Sensitivity Results (additive Gaussian $\sigma$)}
\label{tab:noise}
\begin{tabular}{ccccc}
\hline
\textbf{$\sigma$} & \textbf{Precision} & \textbf{Recall} & \textbf{F1-Score} & \textbf{Seq. Sim.} \\ \hline
0 & 9.4\% & 75.0\% & 16.7\% & 3.0\% \\
5 & 8.8\% & 75.0\% & 15.8\% & 3.1\% \\
10 & 8.6\% & 75.0\% & 15.4\% & 2.8\% \\
20 & 5.7\% & 75.0\% & 10.7\% & 2.1\% \\
30 & 6.8\% & 75.0\% & 12.5\% & 2.7\% \\
\hline
\end{tabular}

\end{table}

\subsection{Proximity Threshold Tuning}
The proximity threshold controls how close in pixels two raw detections must be to
be merged into a single touch event before grid mapping. We sweep it from 10 to
50\,px (Table~\ref{tab:proximity}). The metrics do not move at all: precision,
recall, F1, and sequence similarity are identical across the entire range. This is
itself diagnostic. Because the detector emits a dense, spatially diffuse cloud of
contacts (Section~\ref{sec:audit}), merging neighbours within any of these radii
neither removes the spurious points nor changes which grid cell wins, so the
parameter has no operating range that improves keystroke recovery.

\begin{table}[h]
\centering
\caption{Proximity Threshold Sweep (staged clip). Metrics are invariant to the
threshold, indicating saturation rather than a tunable operating point.}
\label{tab:proximity}
\begin{tabular}{ccccc}
\hline
\textbf{Threshold (px)} & \textbf{Precision} & \textbf{Recall} & \textbf{F1-Score} & \textbf{Seq. Sim.} \\ \hline
10 & 9.4\% & 75.0\% & 16.7\% & 3.0\% \\
20 & 9.4\% & 75.0\% & 16.7\% & 3.0\% \\
30 & 9.4\% & 75.0\% & 16.7\% & 3.0\% \\
40 & 9.4\% & 75.0\% & 16.7\% & 3.0\% \\
50 & 9.4\% & 75.0\% & 16.7\% & 3.0\% \\
\hline
\end{tabular}

\end{table}

\section{Real-World Generalization Audit}
\label{sec:audit}
The experiments above use one staged clip in which the device region is known and
fixed. To test whether the pipeline transfers to uncontrolled footage, we run the
identical detector on \NumRealVideos\ real, publicly licensed third-person phone
videos (Pexels, CC0; listed in \code{data/sources.json}), spanning a touchscreen
dialer, blank-screen and green-screen typing, a rear view, and a rotary phone.
These clips are the kind of over-the-shoulder footage the threat model assumes.
We do not fabricate keystroke labels for them; instead we measure the raw quantity
the detector emits, which needs no ground truth and exposes the failure mechanism
directly.

\begin{table}[h]
\centering
\footnotesize
\caption{Detection volume of the BEHINT pipeline across the staged clip and
\NumRealVideos\ real videos. ``Det./frame'' is the mean number of touch points the
detector emits per frame with a 95\% bootstrap CI; ``Peak'' is the busiest single
frame; ``Skin\%'' is the share emitted by the HSV skin filter.}
\label{tab:audit}
\begin{tabular}{@{}llccc@{}}
\toprule
Video & Content & Det./frame [95\% CI] & Peak & Skin\% \\
\midrule
\code{staged} & Staged (BEHINT clip) & 13.7 [13.1, 14.3] & 35 & 99\% \\
\midrule
\code{dialer} & Touchscreen dialer & 2.3 [1.3, 3.5] & 43 & 39\% \\
\code{white} & Typing, blank screen & 158.7 [155.5, 161.8] & 205 & 98\% \\
\code{greenscr} & Typing, green screen & 41.3 [38.8, 43.9] & 74 & 21\% \\
\code{backview} & Typing, rear view & 91.3 [89.7, 92.9] & 122 & 54\% \\
\code{rotary} & Rotary dial & 56.6 [53.9, 59.4] & 108 & 100\% \\
\bottomrule
\end{tabular}

\end{table}

\begin{figure}[h]
\centering
\includegraphics[width=0.95\linewidth]{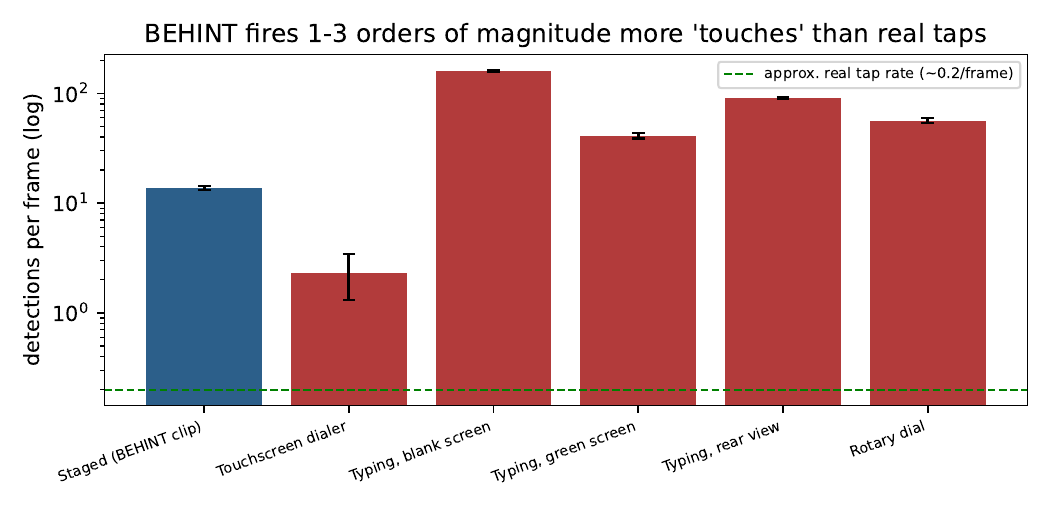}
\caption{Touch detections per frame (log scale) per video. A real typing session
produces on the order of \(0.2\) deliberate taps per frame; the detector emits one
to three orders of magnitude more.}
\label{fig:audit}
\end{figure}

On the real videos the detector emits a median of \RealMedianDPF\ touch points per
frame (range \RealMinDPF--\RealMaxDPF), peaking at \MaxSingleFrame\ points in a
single frame (Table~\ref{tab:audit}, Figure~\ref{fig:audit}). A person typing taps
the screen on the order of a few times per second, well under one deliberate
contact per frame, so the pipeline over-reports contacts by one to three orders of
magnitude. The excess is dominated by the HSV skin filter, which accounts for up to
\SkinMaxFrac\ of the points on some clips because it responds to the entire visible
hand rather than to a fingertip touching glass. This is the mechanism behind the
single-digit precision measured on the staged clip: the system does not detect
\emph{taps}, it detects \emph{skin}, and a keypress is recovered only when a fixed,
hand-tuned device region and a known layout discard almost all of that output. The
audit shows the staged result does not generalize: without per-video calibration
the detector produces a stream of spurious contacts from which no keystroke
sequence can be read.

Table~\ref{tab:methods} attributes each video's detections to the two modalities
that fire (motion and edge feed the same point stream and are not separable
per-point). The HSV skin filter dominates on every clip with a visible bare hand
against a plain background (up to \SkinMaxFrac\ on the rotary and white-screen
clips), and only cedes to MediaPipe on the green-screen and dialer clips where the
chroma background suppresses skin segmentation. The detector's behaviour is thus
governed by skin-coloured area, not by contact, which is precisely the wrong signal
for keystroke recovery.

\begin{table}[h]
\centering
\caption{Share of detections contributed by the HSV skin filter versus MediaPipe
per video. Skin dominates wherever a bare hand is visible against a plain
background.}
\label{tab:methods}
\begin{tabular}{lccc}
\hline
\textbf{Video} & \textbf{Skin \%} & \textbf{MediaPipe \%} & \textbf{Dominant} \\ \hline
Staged & 98.7\% & 1.3\% & Skin \\
Dialer & 38.9\% & 61.1\% & MediaPipe \\
White screen & 98.2\% & 1.8\% & Skin \\
Green screen & 20.8\% & 79.2\% & MediaPipe \\
Rear view & 54.0\% & 46.0\% & Skin \\
Rotary & 100.0\% & 0.0\% & Skin \\
\hline
\end{tabular}

\end{table}

\section{Limitations}
\label{sec:limitations}
The central limitation is ground truth. Only the staged clip has known keystrokes,
so the precision, recall, F1, and sequence-similarity figures are measured on a
single 120-frame sequence; they characterise this pipeline on this clip and should
not be read as a population estimate. We mitigate this in two ways. First, the
real-world audit (Section~\ref{sec:audit}) evaluates generalization on
\NumRealVideos\ additional real videos using a label-free metric, detection volume,
that needs no keystroke annotation and exposes the failure mechanism directly.
Second, every number is regenerated from the result CSVs by \code{gen\_macros.py}
and independently re-checked by \code{verify\_numbers.py}, so the staged-clip
figures are at least exact and reproducible. A quantitative accuracy estimate across
many subjects, devices, and viewing angles would require a labelled multi-video
dataset with frame-level touch annotations, which we identify as the necessary next
step rather than claim here. The ablation, resolution, noise, and proximity sweeps
are likewise computed on the staged clip and describe sensitivity, not field
performance.

\section{Discussion}
\label{sec:discussion}
Our measurements do not support the claim that this class of multi-modal filter pipeline performs virtual keylogging in practice. On the one clip with ground truth, the combined system reaches an F1-score of only \AblationFOneAllCombined\ and a sequence similarity of \AblationSeqAllCombined, far from a reliable operational benchmark for virtual keylogging. The real-world audit (Section~\ref{sec:audit}) explains why: across \NumRealVideos\ real videos the detector emits a median of \RealMedianDPF\ contacts per frame, dominated by a skin filter that fires on the whole hand, so it isolates a keypress only when a fixed device region and a known keypad layout are supplied by hand and discard nearly all of its output. The honest reading is that recovering keystrokes from over-the-shoulder video with classical multi-modal filters does not work outside a calibrated staged setting.

\section{Future Work}
\label{sec:future}
The single most important next step is ground truth at scale. A labelled
multi-video dataset, several subjects typing known passcodes on different devices
and at different viewing angles, with frame-level touch annotations, would replace
the single staged clip and let per-character accuracy, edit distance, and
passcode-recovery rate be measured as population estimates rather than on one
sequence. On the modelling side, the audit points to two concrete replacements. The
skin filter is the dominant source of false contacts and should be replaced by a
learned fingertip-contact classifier that fires on a fingertip touching glass
rather than on skin-coloured area; and the hand-tuned device region should be
replaced by automatic device detection and perspective rectification so the
pipeline can be pointed at uncontrolled footage without per-video calibration.
Only once these are in place would a fair test of whether over-the-shoulder
keystroke inference is feasible at all be possible.

\section{Conclusion}
\label{sec:conclusion}
A training-free multi-modal filter pipeline does not recover keystrokes from
over-the-shoulder video outside a calibrated staged clip. The staged F1-score of
\AblationFOneAllCombined\ collapses on uncontrolled footage, where the detector
emits a median of \RealMedianDPF\ contacts per frame and responds to visible skin
rather than to contact. Reliable virtual keylogging from this class of pipeline
would require learned contact detection and automatic device-region handling, not
the off-the-shelf filters evaluated here.

\bibliographystyle{IEEEtran}
\bibliography{bibliography}

\end{document}